\documentclass[runningheads]{llncs}
\usepackage{graphicx}
\usepackage{amsmath,amssymb} 
\usepackage{color}
\usepackage[width=122mm,left=12mm,paperwidth=146mm,height=193mm,top=12mm,paperheight=217mm]{geometry}
\usepackage{array}
\usepackage{soul}

\newcolumntype{P}[1]{>{\centering\arraybackslash}p{#1}}

\begin{document}
\pagestyle{headings}
\mainmatter
\def\ECCV16SubNumber{137}  

\title{Combining the Best of Convolutional Layers and Recurrent Layers: A Hybrid Network for Semantic Segmentation} 

\titlerunning{A Hybrid Network for Semantic Segmentation}
 
\author{Zhicheng Yan$^1$, Hao Zhang$^2$, Yangqing Jia$^3$, Thomas Breuel$^4$, Yizhou Yu$^{1,5}$}
\institute{$^1$University of Illinois at Urbana-Champaign, $^2$Carnegie Mellon University, \\
$^3$Facebook, $^4$Google, $^5$The University of Hong Kong}

\maketitle
\vspace{-15pt}
\begin{abstract}
State-of-the-art results of semantic segmentation are established by Fully Convolutional neural Networks (FCNs). FCNs rely on cascaded convolutional and pooling layers to gradually enlarge the receptive fields of neurons, resulting in an indirect way of modeling the distant contextual dependence. In this work, we advocate the use of spatially recurrent layers (i.e. ReNet layers) which directly capture global contexts and lead to improved feature representations. We demonstrate the effectiveness of ReNet layers by building a Naive deep ReNet (N-ReNet), which achieves competitive performance on Stanford Background dataset. Furthermore, we integrate ReNet layers with FCNs, and develop a novel Hybrid deep ReNet (H-ReNet). It enjoys a few remarkable properties, including full-image receptive fields, end-to-end training, and efficient network execution. On the PASCAL VOC 2012 benchmark, the H-ReNet improves the results of state-of-the-art approaches Piecewise~\cite{lin2015efficient}, CRFasRNN~\cite{zheng2015conditional} and DeepParsing~\cite{liu2015semantic} by $3.6\%$, $2.3\%$ and $0.2\%$, respectively, and achieves the highest IoUs for 13 out of the 20 object classes\footnotemark.

\end{abstract}

\vspace{-2em}
\section{Introduction}
\vspace{-5pt}
\footnotetext{A part of this work was done when Zhicheng Yan interned at Google Brain.}

Convolutional Neural Networks (CNNs) have achieved notable successes in a variety of visual recognition tasks, such as image classification~\cite{he2015deep,yan2015hd} and object detection~\cite{ren2015faster}.
By replacing fully-connected layers with $1\times 1$ convolutional layers, classification CNNs can be effortlessly transformed into Fully Convolutional Networks (FCNs)~\cite{long2014fully}, which take an input image of arbitrary size and predict a semantic label map.
Though FCNs have achieved state-of-the-art results in semantic segmentation tasks \cite{zheng2015conditional,liu2015semantic}, it suffers a few limitations in modeling distant contextual regions. Specifically, the receptive field of a neuron in the convolutional layer of FCNs usually corresponds to a local area of the input image. However, in semantic segmentation, contextual evidences from distant areas of the image are usually crucial for reasoning and prediction. For example, when labeling the middle area of an image, seeing the pattern of the sea on top of the image and the pattern of a hill at the bottom increases our confidence of making the correct prediction ``beach'', yet the limited size of the local receptive fields inhibits the FCN to capture such spatially long-range dependence across different local areas. Although one can artificially adjust the size of the receptive field to cover the entire image, this implicit modeling usually leads to an ineffective way of encoding long-range context. How the long-range dependence will propagate across filters and whether they will complement each other to help the reasoning still remain unclear. These limitations drive us to seek for a more explicit modeling of global context and long-range dependence.

Recurrent Neural Networks (RNNs) have demonstrated strong capabilities of modeling long-term contextual dependence in speech recognition and language understanding~\cite{graves2013generating,graves2014towards}. In this paper, we introduce the spatially recurrent layer (i.e. ReNet layer) to address the aforementioned limitations. In a ReNet layer, RNNs sweep vertically and horizontally across the image. With gating and memory units to adaptively forget, memorize and expose the memory contents at each running step, the RNN directly propagates spatially long-range information throughout its hidden units, and generates image representations that better capture global context.
We stack ReNet layers to build a naive deep spatially recurrent network (N-ReNet), and achieve comparable performance to state-of-the-art approaches. We visualize the intermediate feature maps produced by N-ReNet, and observe a form of hierarchical feature representations, similar to those generated by deep CNNs. To boost the performance, we further integrate ReNet layers with existing FCNs to form a deep hybrid network (H-ReNet), of which the convolutional layers and pooling layers extract local features, and the recurrent layers perform spatial long-range information propagation. The H-ReNet exhibits a few favorable properties. First, by employing recurrent layers to facilitate the long-range dependence propagation, H-ReNet directly supports full-image receptive fields. Second, incorporating ReNet layers to capture global context improves the learned feature representations, with compelling performance on both region recognition and boundary localization. Third, H-ReNet is end-to-end trainable with efficient forward and backward executions. Its computations in the recurrent layers can be easily parallelized, thus enables H-ReNet to fully exploit the computational power of modern GPUs (compared to the graphical models used in FCN-based models). Overall, recurrent layers substantially improve the results at a negligible increase of computational costs.

To conclude, the contributions of our work are two-fold. First, we introduce spatially recurrent layers (i.e. ReNet layers) for semantic segmentation, and show that by simply stacking ReNet layers, the resulting N-ReNet achieves competitive performance on \textit{Stanford Background} dataset. Second, we construct a hybrid network (i.e. H-ReNet) by appending recurrent layers on top of FCNs. We extensively evaluate the H-ReNet on benchmark PASCAL VOC 2012 with both internal ablation studies and external comparisons. We show that H-ReNet yields improved feature representations over FCNs and achieves state-of-the-art results.

\vspace{-10pt}
\section{Related Work}
\noindent \textbf{Nonparametric Methods}.
Nonparametric methods have achieved remarkable performance in semantic segmentation~\cite{tighe2013finding,Tighe:2010:SSN,LiuYT11,singh2013nonparametric,eigen2012nonparametric}.
The core idea is retrieving similar patches from a database of fully annotated images, and transferring the labels from the annotated images to the query image. Specifically, the query image is matched against the annotated database using both holistic image representations as well as superpixels. Probabilistic graphical models (e.g. MRF, CRF) are then introduced to model the semantic context and obtain a spatially coherent semantic label map~\cite{russell2009associative,kohli2009robust,li2013exploring,ramalingam2008exact}. 
Nonparametric methods divide the segmentation task into individual steps. Each step requires a careful design, and the entire process is not amenable to joint optimization.

\noindent \textbf{Parametric Methods}.
Parametric methods have been dominated by FCN-based models, which can be classified into two lines. In the first line, the FCN takes input of bounding boxes which encompass image regions with high objectness~\cite{zitnick2014edge}, and outputs a segmentation mask for each bounding box. The final segmentation map is obtained by merging individual ones for the bounding boxes \cite{dai2014convolutional,dai2015boxsup,hariharan2014hypercolumns,hariharan2014simultaneous,noh2015learning,girshickregion}. By contrast, in the second line, the whole image is directly fed into the segmentation net, and a complete segmentation mask is generated at once~\cite{pinheiro2013recurrent}. Due to the pooling layers of CNNs, the output mask is usually not sufficiently sharp, and region boundaries are not clearly localized. An additional graphical model layer (e.g. MRFs and CRFs) is thus introduced to capture pixel interactions and respect region boundaries. The graphical model can either be applied as a separate post-processing step~\cite{chen2014semantic} or be plugged into a deep neural net with joint optimization~\cite{zheng2015conditional,schwing2015fully}, 
both at a high cost of extra computations. Besides FCN-based methods, Mostajabi et al.~\cite{mostajabi2014feedforward} propose to label the superpixels using zoom-out features, which include pixel-level, region-level and global features extracted from a deep neural network.

\noindent \textbf{RNNs for Visual Recognition}.
Exploiting recurrent neural networks for visual recognition is an active field of research. Byeon et al. propose a cascaded structure consisting of alternating 2D Long Short Term Memory (LSTM) and convolutional layers, and report comparable results to state-of-the-art on both \textit{Stanford Background} and \textit{SiftFlow} datasets~\cite{Tighe:2010:SSN}. 
Bell et al. develop the IRNN layer for object detection to generate features that are not limited to the bounding box of an object proposal~\cite{bell2015inside}. The recently proposed ReNet architecture is a scalable alternative to CNNs for image recognition~\cite{visin2015renet}. We build our models upon ReNet layers, to capture the global contexts as well as enjoy its property of efficient parallelization. 
In contrast to the IRNN layer where a naive ReLU RNN is implemented,
we employ sophisticated LSTM with various gating units to adaptively forget, memorize and expose the memory contents. 
Empirically, we observe better performance by the ReNet LSTM layer for our task.

\section{Spatially Recurrent Layer Group}
We first introduce the spatially recurrent layer, which is a novel component of our work. Following the naming convention in \cite{visin2015renet}, we refer to a spatially recurrent layer as a ReNet layer. Specifically, a ReNet layer receives either an input image or an input feature map $I$ of size $H \times W $, and divides it into a grid of $h \times w $ patches with patch size $s \times t$ where $h=\left \lceil \frac{H}{s} \right \rceil$ and $w=\left \lceil \frac{W}{t} \right \rceil$. It has two 1D RNNs with independent weights sweeping across the grids vertically (or horizontally) in opposite directions, one in forward and the other in backward. We choose the LSTM unit described in~\cite{zaremba2014recurrent} as our basic RNN implementation because of its outstanding property of overcoming the issue of vanishing gradient, but note that other RNN variants, such as Recurrent Gated Units~\cite{chung2015gated}, might be also suitable. An 1D LSTM unit includes the forget gate $\mathbf{f_t}$, the input gate $\mathbf{i_t}$, the cell input $\mathbf{\widetilde{C}_t}$, the cell memory $\mathbf{C_t}$, the output gate $\mathbf{o_t}$ and the hidden state $\mathbf{h_t}$, which enable the LSTM unit to adaptively forget, memorize and expose its memory content $\mathbf{C_t}$ at each running step $t$.


Formally, the ReNet layer takes a 2D map as input, sweeps across the grids in opposite directions, and updates the cell memory $\mathbf{C_{y,x}}$ and the hidden state $\mathbf{h_{y,x}}$ of its LSTM units at pixel positions $(y,x)$ as

\vspace{-0.5em}
\begin{equation}
\label{eqn:renet_lstm}
\begin{split}
(\mathbf{h^F_{y,x}},\mathbf{C^F_{y,x}})=LSTM^F(\mathbf{I_{y,x}},\mathbf{h^F_{y-1,x}},\mathbf{C^F_{y-1,x}})   \mbox{     for  } y=1,..,H \\
(\mathbf{h^B_{y,x}},\mathbf{C^B_{y,x}})=LSTM^B(\mathbf{I_{y,x}},\mathbf{h^B_{y+1,x}},\mathbf{C^B_{y+1,x}})    \mbox{     for  } y=H,...,1
\end{split}
\end{equation}

\noindent where we use the the superscripts $F$ and $B$ to denote forward and backward directions.
As the scanning of two LSTMs is independent of each other, we can easily parallelize their computations for additional speedup. By concatenating hidden states $\mathbf{C^F_{y,x}}$ and $\mathbf{C^B_{y,x}}$, each of which has $d$ hidden units, we can obtain a composite feature map of size $h \times w \times 2d$, with a receptive field comprised of all the patches within the same column. 
By stacking two ReNet layers with orthogonal sweeping directions (e.g. horizontal and vertical), we can obtain an output feature map fully covering the input image, as shown in Figure 1.
We refer to two ReNet layers with orthogonal sweeping directions as a \textit{recurrent layer group}. 

We construct a Naive deep ReNet (N-ReNet) by stacking multiple layer groups on top of each other. The first group takes raw pixels as input and the output of the last group is passed through a softmax layer to produce dense predictions. To align the channel number of output feature maps with the number of semantic labels, we append an auxiliary $1\times 1$ convolutional layer on top of the last ReNet layer, with the number of convolutional kernels identical to the number of labels.

Similarly, we can further integrate the ReNet layers with FCNs by appending a recurrent layer group at the end of a pretrained FCN, such as VGG-16 net~\cite{simonyan2014very} and GoogleNet~\cite{szegedy2014going} pretrained on ImageNet for image classification. It enables us to exploit both the advantages of FCNs in capturing local context and the capability of ReNet layers in modeling distant and global context. We refer to this hybrid network as an \textit{H-ReNet}. Both \textit{N-ReNet} and \textit{H-ReNet} will be evaluated in our experiments.

\vspace{-1em}
\section{Experiments}
We first train a N-ReNet from scratch and evaluate it on the \textit{Stanford Background} dataset. We compare the N-ReNet to other competing methods that also do not use pretrained model. Our focus in this part is not establishing new state-of-the-art results. Instead, we aim to demonstrate the effectiveness of the proposed recurrent layer group. 
After confirming the capability of the recurrent layer group, we then conduct experiments for the H-ReNet, where ReNet layers are appended at the end of a pretrained FCN. We evaluate the H-ReNet on \textit{PASCAL VOC 2012} (\textit{VOC12}) dataset.

We implement our models based on Caffe \cite{Jia13caffe}. All experiments are conducted on a single NVIDIA K40c GPU. In particular, the ReNet LSTM layer has been efficiently implemented on GPU. The vertical and horizontal sweepings in a recurrent layer group are parallelized to improve efficiency.

\noindent \textbf{Training}. Both N-ReNet and H-ReNet require a minimal input image size of $H_{min} \times W_{\min}$. Reflection padding is  applied if an input image is smaller than $H_{min} \times W_{\min}$. At the training stage, randomly cropped patches of size $H_{min}\times W_{min}$ with random horizontal flipping are fed into the model in a mini-batch size of 10. We set $(H_{min}, W_{\min})$ to be $(240,320)$ and $(400,500)$ on two datasets \textit{Stanford Background} and \textit{VOC12}, respectively.

Since ReNet layers are differentiable, both N-ReNet and H-ReNet are end-to-end trainable by stochastic gradient descent. Specifically, we employ the pixelwise multinomial cross entropy \cite{long2014fully} with equal weights for all semantic labels as the loss function, and train the model parameters by standard backpropagation.

\noindent \textbf{Testing}. At the testing stage, the network can take an image at its original size, as both convolutional layers and ReNet LSTM layers can handle inputs of variable size. It produces dense predictions at the original resolution of the test image.

\vspace{-1em}
\begin{figure}[h]
\begin{center}
\includegraphics[width=0.6\textwidth]{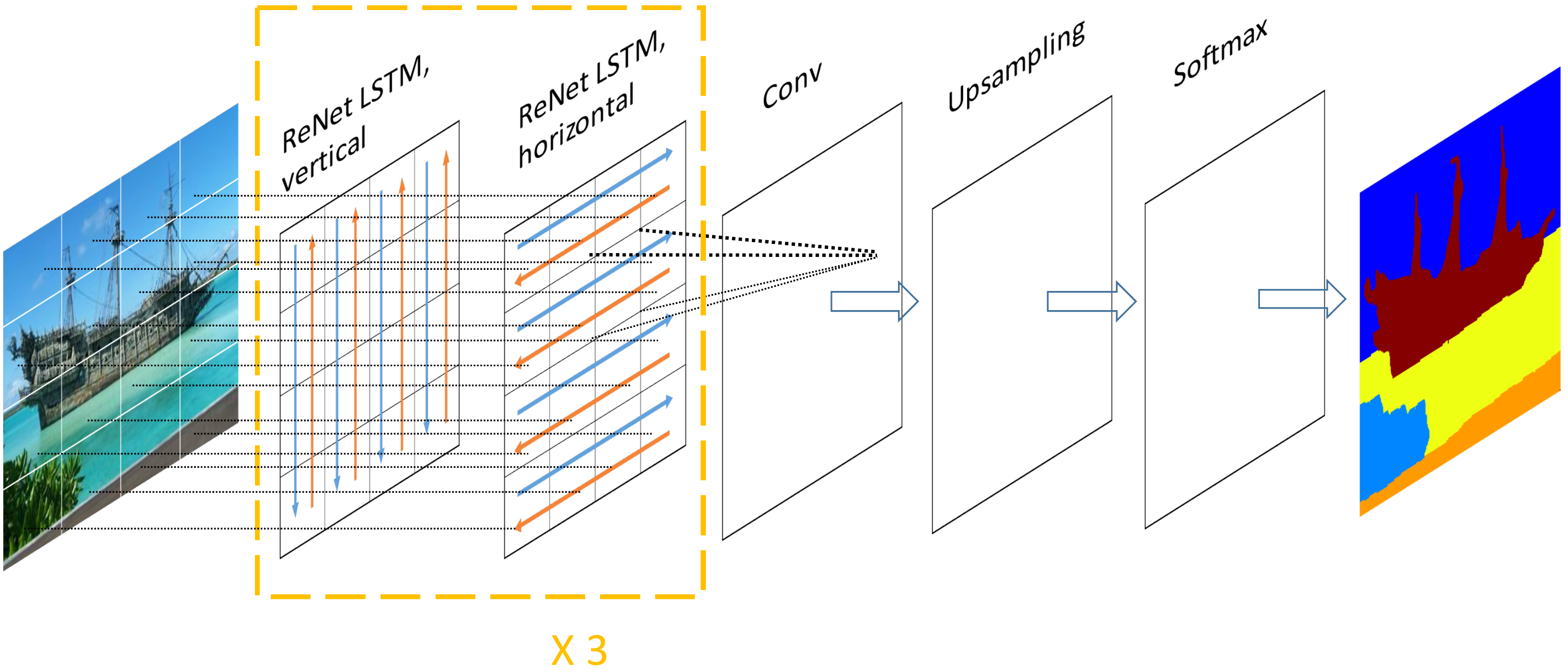}
\end{center}
\vspace{-1em}
\caption{A N-ReNet with three recurrent layer groups.}
\label{fig:stanford_net}
\vspace{-2em}
\end{figure}

\subsection{Evaluations of N-ReNet}
We evaluate the N-ReNet on \textit{Stanford Background} dataset, which contains 715 images of outdoor scenes with 8 labels. We randomly and evenly divide the images into 5 sets, and report 5-fold cross validation results. 

\begin{figure}[]
\begin{center}
\includegraphics[width=.9\linewidth]{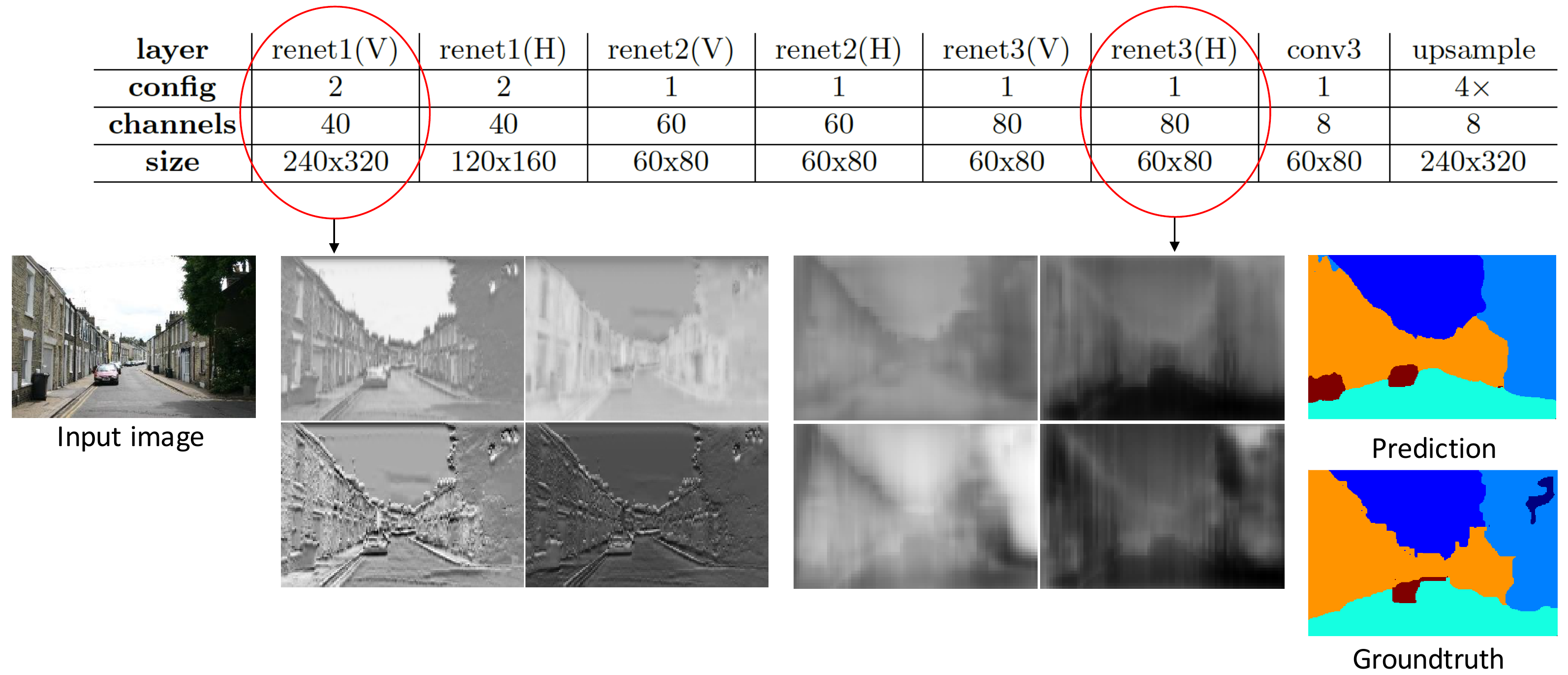}
\end{center}
\vspace{-2em}
   \caption{A  N-ReNet with three recurrent layer groups. The first 4 feature maps of the layers \textit{renet1(V)} and \textit{renet3(H)} are visualized as grayscale images, where we use H (horizontal) and V (vertical) to denote the sweeping direction of the ReNet layer.}
\label{fig:renet_3_lay_stanford}
\vspace{-2em}
\end{figure}

We first configure a N-ReNet 
consisting of 3 ReNet layer groups with increasing numbers of neurons, as shown in Figure \ref{fig:renet_3_lay_stanford}. The first layer group sweeps over $2\times 2$ patches while all other groups scan $1\times 1$ patches (i.e. pixels), thereby the spatial resolution is reduced by a factor of 4. An auxiliary $1\times 1$ convolutional layer with 8 kernels is placed  on top of  ReNet layers for the purpose of aligning the  number of feature maps with the number of semantic labels. Finally, an upsampling layer is appended to restore the label map to the original spatial resolution via bilinear interpolation. It is  noticeable that neurons in the last convolutional layer has a receptive field covering the entire input image.


In the rest of this section, we first show that the N-ReNet learns hierarchical feature representations similar to those learned by deep CNNs. Then we demonstrate that the N-ReNet achieves comparable accuracy on the Stanford Background dataset with outstanding running efficiency. 
\subsubsection{Hierarchical Feature Representations}
Deep CNNs tend to learn hierarchical feature representations~\cite{zeiler2014visualizing}. It is intriguing to figure out whether the N-ReNet has similar capabilities. In Figure \ref{fig:renet_3_lay_stanford}, we visualize the first 4 feature maps from both an early ReNet layer, \textit{renet1(V)}, and a later ReNet layer, \textit{renet3(H)}. In its early layer, the N-ReNet extracts low-level features that still preserve fine-scale image details (e.g. windows and doors of the buildings). As such details  become less important for semantic region inference, the N-ReNet learns to gradually smooth out them and transform them into high-level discriminative features in  its deep layer, resulting in a form of hierarchical feature representations.

\subsubsection{Results on Stanford Background}
Our N-ReNet, which is comprised of three recurrent layer groups, achieves competitive results on \textit{Stanford Background}, as shown in Table \ref{tab: stanford_accuracy}. We compare the N-ReNet to other methods from three classes. 
First, the N-ReNet improves the pixel accuracy and class accuracy of all nonparametric methods \cite{singh2013nonparametric,Tighe:2010:SSN,eigen2012nonparametric} by at least $2.9\%$ and $5.3\%$ while executes at least $57\times$ faster with an efficient GPU implementation. Second, compared to those built on top of CNNs,
the N-ReNet outperforms the methods in \cite{farabet2012scene,pinheiro2013recurrent,kekecc2014contextually}. Recursive context propagation~\cite{sharma2014recursive} achieves higher labeling accuracy than ours. They  employ multiple CNNs with tied weights to extract multi-scale local features, while the N-ReNet only uses single-scale features. It is noticeable that they rely on superpixels to reduce the computational cost during context propagation, while the superpixel segmentation turns out to be its computational bottleneck, accounting for $0.3$ second out of a total of $0.37$ second. The zoom-out approach in \cite{mostajabi2014feedforward} achieves the highest pixel accuracy by combining both hand-crafted and CNN-based superpixel features. 
Although the computational cost is not reported in \cite{mostajabi2014feedforward}, this method is likely to be computationally expensive as it involves superpixel segmentation, heavy feature extraction (e.g. pixel-level, region-level and global features) and a multi-layer perceptron. Overall, CNN-based methods run at least $5\times$ slower than our N-ReNet.  
Third, compared with another network built on the basis of 2D LSTMs~\cite{byeon2015scene}, N-ReNet improves the pixel and class accuracy by $1.8\%$ and $3.0\%$, respectively. The 2D LSTM in \cite{byeon2015scene} follows the sequential scan-line order and cannot be easily accelerated on GPU. On the contrary, with parallel 1D LSTM computations, the N-ReNet runs more than $18\times$ faster. Overall, our evaluations confirm the effectiveness of the ReNet layer as a novel alternative of FCNs for semantic segmentation. \\

\vspace{-1em}
\begin{table}[h!]
\begin{center}
\resizebox{.7\textwidth}{!}{
    \begin{tabular}{ P{4.9cm} |P{2.0cm} | P{2.2cm}}
    Method & Accuracy($\%$)& Time (sec.)\\ 
    \hline \hline
    Nonparametric parsing \cite{singh2013nonparametric} & 74.1, 62.2 & 20 (CPU)\\ \hline
    Superparsing \cite{Tighe:2010:SSN} & 77.5, N/A & 4.0 (CPU) \\ \hline
    Nonparametric parsing II \cite{eigen2012nonparametric} & 75.3, 66.5 & 16.6 (CPU) \\ \hline    
    \hline
    Single-scale ConvNet \cite{farabet2012scene} & 66.0,56.5 & 0.35 (CPU)\\ \hline
    Multi-scale ConvNet \cite{farabet2012scene} & 78.8,72.4 & 0.6 (CPU)\\ \hline    
    Recurrent CNN \cite{pinheiro2013recurrent} & 76.2,67.2 & 1.1 (CPU) \\ \hline   
    Multi-CNN + rCPN Fast \cite{sharma2014recursive}& 80.9,\textbf{78.8} & 0.37(GPU) \\ \hline  
    Augmented CNN\cite{kekecc2014contextually} & 76.4, 68.5 & N/A \\ \hline
    Zoom-out\cite{mostajabi2014feedforward} & \textbf{82.1},77.3  & N/A \\ \hline \hline
    Deep 2D LSTM\cite{byeon2015scene} &  78.6,68.8 & 1.3 (CPU) \\ \hline 
    \hline
    N-ReNet &  80.4,71.8 & \textbf{0.07 (GPU)} \\   
    \hline
    \end{tabular}
}
\end{center}
\caption{Comparisons of pixel/class accuracy and testing time. For fair comparisons, the results of Multiscale ConvNet  \cite{farabet2012scene} are from the variant without the sophisticated gPb hierarchical segmentation~\cite{arbelaez2011contour}.  
}
\label{tab: stanford_accuracy}
\vspace{-2em}
\end{table}

\vspace{-2em}
\subsection{Evaluations of H-ReNet}
Given the evidence that the ReNet layer alone is effective for inferring semantic regions, we are interested in answering a more ambitious question. Can we improve the feature representations of FCNs by adding a ReNet layer group? To answer, we build an H-ReNet consisting of convolutional, pooling and ReNet layers, and evaluate it on the \textit{VOC12}. We use the augmented dataset from \cite{hariharan2011semantic} with $10,582$, $1,449$ and $1,456$ images in the \textit{training} set, \textit{validation} set and \textit{test} set, respectively. We evaluate on the \textit{validation} set when only the \textit{training} set is used for training. We also report results on the \textit{test} set when both the \textit{training} set and the \textit{validation} set are used for training. We follow the \textit{comp6} evaluation protocol and employ mean Intersection over Union (IoU) as our evaluation metric~\cite{everingham2010pascal}. 

We also consider the setting that uses extra annotations from Microsoft COCO dataset \cite{lin2014microsoft}. This dataset contains $123,287$ images in $80$ semantic labels, among which $96,685$ images have overlapped labels with \textit{VOC12} and more than $900$ pixels annotated by these labels in the image. 

In the following we introduce the model structures we use for the experiments, including a baseline FCN architecture adapted from pretrained classification CNNs, and our H-ReNet architecture with multiple variants.
\vspace{-1em}
\subsubsection{Baseline FCN Architecture}
We follow the practice in \cite{chen2014semantic} and adapt the pretrained VGG-16 network into a baseline FCN with the architecture shown in Table \ref{tab: baseline}. Specifically, we reuse the first 10 layers of VGG-16, which include 5 groups of convolutional layers (\textit{conv1}-\textit{conv5}) and 5 max-pooling layers (\textit{pool1}-\textit{pool5}). Each max-pooling layer reduces the size of the feature map by a half. Overall, the size of the feature map is reduced by a factor of 32. Although reducing the resolution of feature maps may not compromise the accuracy in image classification tasks, high-resolution feature maps are important in semantic segmentation to reconstruct accurate region boundaries in the final semantic map~\cite{long2014fully}.
Therefore, we reduce the strides of the layers \textit{pool4} and \textit{pool5} from 2 to 1, so that the overall downsampling factor decreases from 32 to 8. This technique is known as \textit{hole algorithm} in \cite{chen2014semantic} and \textit{dilated convolution} in \cite{yu2015multi}.


\begin{table}[t]
\small
\begin{center}
\resizebox{1.0\textwidth}{!}{
    \begin{tabular}{ P{1.5cm} P{1.0cm}  P{0.9cm} P{1.0cm}  P{0.9cm}  P{1.0cm} P{0.9cm} P{1.0cm} P{.9cm} P{1.0cm} P{.9cm} P{1.0cm} P{.9cm} P{.9cm} P{1.2cm}} 
    \multicolumn{15}{c}{\textbf{Baseline FCN}}\\ 
    \hline   
    \textbf{layer} & conv1 & pool1 & conv2 & pool2 & conv3 & pool3 & conv4 & pool4 & conv5 & pool5 & conv6 & conv7 & conv8 & upsample\\ \hline
    \textbf{type} & conv$\times$2 & max & conv$\times$2 & max & conv$\times$3 &max& conv$\times$3 & max & conv$\times$3 & max & conv & conv & conv & upsample \\ \hline  
    \textbf{config} & 3,1,1 & 2,2 & 3,1,1 & 2,2 & 3,1,1 & 2,2 & 3,1,1 & 3,1 & 3,1,2 & 3,1 & 3,1,12 & 3,1,1 & 1,1,1 & 8x\\ \hline
    \textbf{channels} & 64 & 64 & 128 & 128 & 256 & 256 & 512 & 512 & 512 & 512 & 1024 & 1024 & 21 & 21   \\ \hline
    \textbf{activation}  & ReLu & idn & ReLu & idn & ReLu & idn & ReLu & idn & ReLu & idn & ReLu & ReLu & idx & SMAX   \\
\hline
    \end{tabular}
}

\end{center}
\caption{The architecture of the baseline FCN. 
For layer configurations (the 3rd row), we report (\textbf{kernel size}, \textbf{stride}, \textbf{dilation}) for convolutional layers, (\textbf{kernel size}, \textbf{stride}) for max-pooling layers, and (\textbf{upsampling factor}) for bilinear upsampling layers. Notations: \textbf{idn} denotes the identity activation function.}
\label{tab: baseline}
\end{table}

\begin{table}[t]
\begin{center}
\resizebox{.7\textwidth}{!}{
    \begin{tabular}{ P{1.5cm} P{5.0cm}  P{1.4cm}  P{0.9cm}  P{1.3cm} } 
    
    \multicolumn{5}{c}{\textbf{H-ReNet}}\\ \hline
    \textbf{layer ID} & baseline FCN conv1-conv7 & renet1 & conv8 & upsample \\ \hline
    \textbf{type} & - &  ReNet$\times$2 & conv & upsample \\ \hline      
    \textbf{config} & - & 1 & 1,1,1 & $8\times$ \\ \hline
    \textbf{channels} & -  & 240 & 21 & 21 \\ \hline
    \textbf{activation}  & - & idn & idn & SMAX   \\ 
\hline
    \end{tabular}
}
\end{center}
\caption{The architecture of the H-ReNet. For layer configurations (the 3rd row), we report (\textbf{scanning patch size}) for ReNet layers.}
\label{tab: hybrid_net}
\vspace{-2em}
\end{table}

We further append 3 convolutional layers (\textit{conv6}, \textit{conv7} and \textit{conv8}) on top of the last max-pooling layer \textit{pool5}. They play similar roles as the three fully-connected layers in VGG-16 net.
To increase the size of receptive fields, we dilate the convolutional kernels in layer \textit{conv6} to have a stride of 12 \cite{yu2015multi}, which leads to a receptive field of $604\times 604$ with padding. 
Finally, the spatial resolution is restored by upsampling the feature maps from layer \textit{conv8} using bilinear interpolation. Note our baseline FCN is akin to the DeepLab-LargeFOV in \cite{chen2014semantic}, but achieves higher accuracy ($63.4\%$ v.s. $62.3\%$).

\vspace{-1.5em}
\subsubsection{H-ReNet Architecture}


The architecture of the H-ReNet is derived by inserting a recurrent layer group into the baseline FCN. An example of an H-ReNet is shown in Table \ref{tab: hybrid_net}. We insert one ReNet layer group \textit{renet1}, which includes two ReNet layers with orthogonal scanning directions, between layers \textit{conv7} and \textit{conv8}. Compared to the baseline FCN where the input to \textit{conv8} is the output of another convluational layer \textit{conv7}, in the H-ReNet, the input to \textit{conv8} is from a ReNet layer group. This provides an ideal testbed for us to verify the effectiveness of the ReNet layer group.  
\vspace{-1.5em}
\subsubsection{Multi-layer Feature Combination} 
\begin{figure}[t]
\begin{center}
\includegraphics[width=0.75\textwidth]{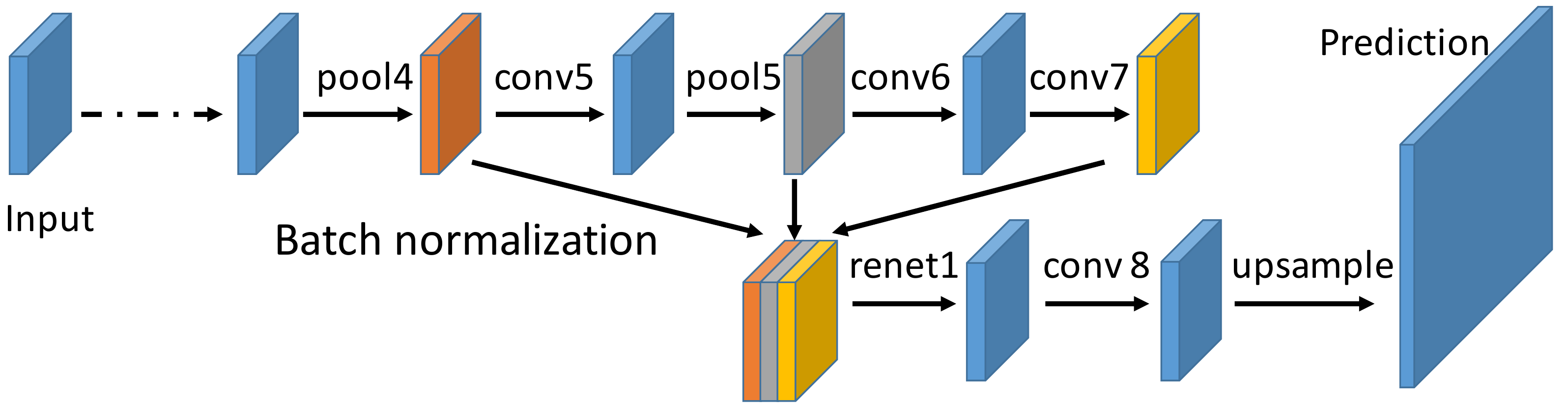}
\end{center}
\vspace{-1em}
\caption{Multi-layer feature concatenation with batch normalization. }
\label{fig:hybrid_net_ftr_concat}
\vspace{-1.5em}
\end{figure}

Combining features from multiple layers yields more discriminative features for semantic segmentation and  object detection~\cite{hariharan2014hypercolumns,bell2015inside}. We therefore concatenate feature maps from \textit{pool4}, \textit{pool5} and \textit{conv7}, as shown in Figure \ref{fig:hybrid_net_ftr_concat}. As the magnitude of the feature values from different layers may vary substantially, we further normalize the feature maps before the concatenation in case the final combination is dominated by one or two feature maps.
We experiment with both L2 normalization and batch normalization~\cite{ioffe2015batch}. Assume that the size of the feature map is $N_f \times C_f \times H_f \times  W_f$ where $N_f,C_f,H_f$ and $W_f$ denote minibatch size, channel number, height and width, respectively. For L2 normalization, we normalize each map of size $C_f \times H_f \times  W_f$ to have unit L2 norm and then uniformly scale it by a constant $\lambda$, chosen by a grid search from $\{100,1000,10000\}$ on a held-out set. For batch normalization, we collect a number of $N_f \times H_f \times W_f$ samples and normalize them to have zero mean and unit variance. 

In the rest part of this section, we first introduce the experiment settings we use to evaluate the H-ReNet. Then, we conduct ablation studies on the VOC12 dataset by comparing the baseline FCN, the H-ReNet and its variants. Finally, we compare the H-ReNet to other start-of-the-art methods, and demonstrate its superiority.
\vspace{-1.5em}
\subsubsection{Experimental Settings}
For both the baseline FCN and the H-ReNet, the parameters from layer \textit{conv1} to layer \textit{conv5} are initialized from VGG-16 pretrained model. Parameters in other convolutional layers are randomly initialized by sampling from a Gaussian distribution ($\mu=0$, $\sigma = 0.1$). The parameters of ReNet layers are randomly initialized by sampling from a uniform distribution over $[-0.2,0.2]$. 
We empirically observe that a full-pass backpropagation is unnecessary during training. Skipping layers \textit{conv1} and \textit{conv2} during backpropagation has negligible impact on the final results but can speeds up the network training. 
For the baseline FCN, we find $12,000$ training iterations are sufficient. For the H-ReNet, we train for $30,000$ iterations when the multi-layer feature combination is disabled, and $40,000$ iterations when enabled. The initial learning rates for both models are set to $0.001$ and decreased by a factor of 10 in the middle of the training.


With extra training data from Microsoft COCO are added, we first pretrain the H-ReNet for $120,000$ iterations, using a combination of the \textit{training} set and the \textit{validation} set from both \textit{VOC12} and Microsoft COCO. The initial learning rate $0.001$ is decreased by a factor of 10 for every $50,000$ iterations. As the annotation masks from COCO dataset are coarser than those from \textit{VOC12}, finetuning on \textit{VOC12} data for another $30,000$ iterations is necessary. Evaluation is performed on the \textit{test} set. 


\vspace{-1em}
\subsubsection{Results on PASCAL VOC 2012}
We compare H-ReNet and its variants to the baseline FCN on the \textit{validation} set of VOC12 in Table \ref{tab: seg_internal_comp_baseline}. The baseline FCN achieves a mean IoU $63.4\%$. By inserting one recurrent layer group between \textit{conv7} and \textit{conv8}, with the rest of the structure unchanged, we obtain a substantial improvement of $6.6\%$ on the mean IoU ($70.0\%$ v.s. $63.4\%$). The improvement is completely attributed to the recurrent layer group, which uses two ReNet layers with orthogonal sweeping directions to capture distant context across different local areas. By adding batch normalization~\cite{ioffe2015batch}, we further improve the mean IoU by $0.4\%$ ($70.4\%$ v.s. $70.0\%$). This demonstrates that the ReNet layer can also benefit from normalization techniques. If we further concatenate feature maps from \textit{pool4}, \textit{pool5} and \textit{conv7} before feeding into \textit{renet1}, we further improve the mean IoU by $0.7\%$ ($71.1\%$ Vs $70.4\%$), which demonstrates that ReNet layers can also be complemented by multi-scale feature combination~\cite{hariharan2014hypercolumns}. 
We also experiment with L2 normalization for feature concatenation, but observe no improvement.

\vspace{-2em}
\begin{table}[h]
\begin{center}
\resizebox{.7\textwidth}{!}{
    \begin{tabular}{ P{3.0cm} | P{.8cm} P{1.cm} P{1.7cm} |  P{1.4cm}}
    Variant & BN & MLFB & DenseCRF & IoU($\%$) \\ 
    \hline \hline
    (a) baseline & & &  & 63.4\\ \hline
    (b) H-ReNet &  & & & 70.0\\ \hline
    (c) H-ReNet & \checkmark &  &  & 70.4  \\ \hline
    (d) H-ReNet &  \checkmark &  \checkmark &  & 71.1 \\ \hline \hline
    
    (e) baseline & & &  \checkmark & 67.5 \\ \hline
    (f) H-ReNet &  \checkmark & &  \checkmark & 71.9  \\ \hline
    (g) H-ReNet &  \checkmark & \checkmark &  \checkmark & 72.6  \\
    \hline
    \end{tabular}
}    
\end{center}
\caption{Comparisons between the baseline FCN and the H-ReNet on the validation set of \textit{VOC12}. Notations: \textbf{BN} = Batch Normalization, \textbf{MLFB} = Multi-Layer Feature Combination.}
\label{tab: seg_internal_comp_baseline}
\vspace{-4em}
\end{table}

\subsubsection{Improving region recognition and boundary localization}
\begin{figure}[t!]
\begin{center}
 \centerline{\includegraphics[width=1.0\linewidth]{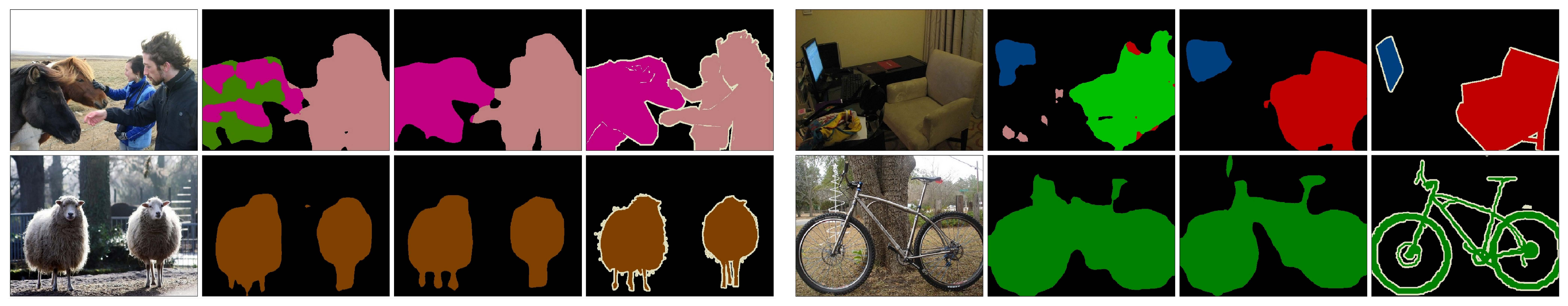}}
\end{center}
\vspace{-1em}
\caption{Qualitative comparisons between the baseline FCN and the H-ReNet. For each example, results of input, baseline, H-ReNet w/o \textit{DenseCRF} and groundtruth are shown from left to right.}
\label{fig:comp_hybrid_baseline_crf}
\vspace{-1em}
\end{figure}

To see how the H-ReNet improves the region recognition and bound localization, we perform qualitative comparisons in Figure \ref{fig:comp_hybrid_baseline_crf}. In the top-left example, the H-ReNet produces a more coherent prediction than the baseline FCN, and in the top-right example the H-ReNet successfully recognize the chair while the baseline FCN fails. In both examples, global context is critical for the model to correctly recognize and separate out the object. 
The baseline FCN has to cascade convolutional and pooling layers to progressively increase the size of its receptive fields, while how well the global context could be encoded by this implicit way of modeling is still an open question.
By contrast, the H-ReNet encodes global context by employing ReNet layers to explicitly propagate information across the entire image. In both models, the spatial resolution of feature maps is greatly reduced. However, compared to the baseline FCN, the H-ReNet can better localize and restore the boundaries, as evidenced in the bottom two examples.

\vspace{-1em}
\subsubsection{LSTM v.s. IRNN} The RNN unit in a ReNet layer can also be instantiated by the IRNN layer~\cite{bell2015inside}. Compared to LSTM units, IRNN units lack gating functions and memory units to overcome the vanishing gradient issue, and it heavily relies on identity initialization to support dependence propagation. We implement an IRNN-based ReNet layer with same number of parameters and compare it to our original H-ReNet. It achieves a mean IoU $67.2\%$, which is $3.2\%$ lower than the results of H-ReNet with 1D LSTM units.

\vspace{-1em}
\subsubsection{Multi-Layer Feature Combination} 
Before concatenating features from different layers, we need to first figure out a set of layers with high-quality and complementary features. To investigate the ``quality'' of the features from each individual layer, 
we build an H-ReNet, where the recurrent layer group are fed feature maps solely from that layer under investigation. For example, to investigate the feature map from \textit{pool4}, we connect \textit{pool4}, \textit{renet1} and \textit{conv8}, and discard the intermediate layers (\textit{conv5} to \textit{conv7}), to obtain an H-ReNet. 
We choose a candidate set of layers $\{\textit{pool4}, \textit{pool5}, \textit{conv7}\}$ for investigation. 
As shown in the left side of Table \ref{tab:comp_hg_input}, the feature maps from deeper layers give better results than those from shallower layers, as deeper layers generally extract high-level features more helpful for reasoning. We are also interested in whether features from different layers are complementary to each other. As shown in the right side of Table \ref{tab:comp_hg_input}, as we gradually add more feature maps from lower layers \textit{pool5} and \textit{pool4}, the accuracy is improved by $0.7\%$ ($71.1\%$ v.s. $70.4\%$). However, the improvement diminishes as more feature maps from other lower layers are concatenated.

\vspace{-1.em}
\begin{table}[t]
\begin{minipage}{.5\linewidth}
    \centering
    \begin{tabular}{ P{3.0cm} | P{1.3cm}}
    ReNet Layer Input & IoU($\%$) \\ 
    \hline \hline
    pool4 & 59.2\\ \hline
    pool5 & 68.3\\ \hline
    conv7 & 70.4 \\ \hline
    \end{tabular}
\end{minipage}
\begin{minipage}{.5\linewidth}
    \centering
    \begin{tabular}{ P{1.0cm} P{1.0cm} P{1.0cm} | P{1.3cm}}
     conv7 & pool5 & pool4 & IoU($\%$) \\ 
    \hline \hline
     \checkmark & & & 70.4\\ \hline
     \checkmark & \checkmark & & 70.9\\ \hline
     \checkmark & \checkmark & \checkmark & 71.1 \\ \hline
    \end{tabular}
\end{minipage}
\vspace{0.5em}
\caption{\textbf{Left}: comparisons of results when ReNet layer takes various inputs. \textbf{Right}: comparisons of multi-level feature combination.}
\label{tab:comp_hg_input}
\vspace{-3.em}
\end{table}


\subsubsection{Training Image Size}  
We experiment with different sizes $(H_{min}, W_{min})$ of training images where $(H_{min},W_{min}) \in$ $\{(256,320)$, $(320,400)$, $(400,500) \}$. For the baseline FCN, the mean IoUs are $63.6\%$, $63.4\%$ and $63.4\%$, respectively. We conclude that FCNs are not sensitive to the input size. A plausible explanation is that the effective size of receptive fields in the last convolutional layer of FCNs is no larger than the smallest option $(256,320)$, thus enlarging the training cropping may not be helpful for the FCN to learn better feature representations. On the other side, the H-ReNet benefits from large training croppings, and achieves mean IoUs of $67.3\%$, $69.3\%$ and $71.1\%$, respectively. 
When a larger training image is fed into the H-ReNet, the ReNet layer has to scan more image patches during every single pass, and the contextual information is also propagated over a longer distance, resulting in better feature representations. 

\vspace{-1.5em}
\subsubsection{Incorporating CRFs} 
While the H-ReNet can improve boundary localization over FCNs, it still produces blurry boundaries due to the reduced resolution of feature maps. To refine the boundaries, we adopt the post-processing step \textit{DenseCRF} in \cite{chen2014semantic}, i.e. we add a fully-connected CRF layer which could be solved by the mean-field algorithm. The CRF consists of both unary and pairwise potentials, where the unary potential is defined as negative log-probability predicted by the network and the pairwise potential is a composition of a spatial kernel and a bilateral kernel. We implement the \textit{DenseCRF} on GPU, and our experiments show that it can improve the results of both the baseline FCN and the H-ReNet by $4.1\%$ and $1.5\%$, respectively (Table \ref{tab: seg_internal_comp_baseline}). The \textit{DenseCRF} models the interactions between every pair of pixels by taking input of low-level features (e.g. pixel position, intensity). 
On the contrary, ReNet layers use high-level features from convolutional layers to model the contextual dependence between different local areas of the image. Empirically, we observe post-processing with CRFs complements ReNet layers for dependence propagation.



\vspace{-1.em}
\subsubsection{Comparisons with Other Approaches}
In Table \ref{tab: speed_comp_voc12}, we compare the H-ReNet with other methods on the \textit{validation} set of VOC12. DeepLab-LargeFOV, with the same base FCN as ours, runs slightly faster but underperforms the H-ReNet by $8.8\%$ in IoU. \textit{DeepParsing}~\cite{liu2015semantic} network underperforms the H-ReNet by $3.3\%$ in IoU and running $25\%$ slower. The \textit{Piecewise} network~\cite{lin2015efficient} slightly outperforms the H-ReNet by $0.8\%$, with a cost of $15\times$ more prediction time. By comparing the H-ReNet to our baseline FCN, we find that adding one recurrent layer group increases the prediction time merely by 0.03 second (0.24 v.s. 0.21). Incorporating \textit{DenseCRF} into the H-ReNet further improves IoU by $1.5\%$ with $2\times$ execution time. Therefore, \textit{DenseCRF} post processing represents a trade-off between segmentation accuracy and execution efficiency.

In Table \ref{tab: results_on_voc}, we compare the H-ReNet with other approaches on \textit{test} set of VOC12.
The \textit{FCN-8s} net~\cite{long2014fully} exploits multi-scale features from various layers by combining predictions from different skip connections. In contrast, we concatenate features from different layers and use them to generate a single prediction. We improves \textit{FCN-8s} by $12.1\%$ ($74.3\%$ Vs. $62.2\%$). The \textit{Zoomed-out} approach~\cite{mostajabi2014feedforward} extracts both hand-crafted and CNN-based features in multiple levels. Those features are concatenated to feed into a MLP network to produce superpixel-wise predictions. The H-ReNet does not rely on superpixel prior, but still outperforms \textit{Zoomed-out} by $5.6\%$ ($74.3\%$ v.s. $69.6\%$).
The \textit{CRFasRNN}~\cite{zheng2015conditional} approach appends recurrent layers with mean-field solvers after the \textit{FCN-8s} net, and performs end-to-end training for all layers. 
Both \textit{Piecewise}~\cite{lin2015efficient} and \textit{DeepParsing}~\cite{liu2015semantic} incorporate more sophisticated graphical models into an FCN, and perform end-to-end training. The H-ReNet alone outperforms all the aforementioned approaches except \textit{DeepParsing}. With the \textit{DenseCRF} integration, it slightly improves the result of \textit{DeepParsing} by $0.2\%$, and establishes new state-of-the-art results. The H-ReNet achieves the highest IoUs on 13 out of 20 object classes. For the \textit{bike} class, \textit{DeepParsing} attains substantially better results ($59.4\%$ v.s. $39.6\%$) than ours by correctly predicting the interior regions of bike wheels as background while the H-ReNet fails to do that. These improvements are attributed to the modeling of high-order label relations by \textit{DeepParsing}. Note that our H-ReNet can also incorporate such sophisticated graphical models introduced in \textit{DeepParsing} and \textit{CRFasRNN}, and jointly optimize the model parameters with preceding layers. We expect this can further improve the results of the H-ReNet, and leave the integration of trainable CRF-based post processing with the H-ReNet as a future work.

\begin{table}[t!]
\begin{center}
\resizebox{.7\textwidth}{!} {
    \begin{tabular}{ P{6.cm} | P{2.9cm} | P{0.8cm}}
    Method & Time (sec.) & IoU($\%$) \\ 
    \hline \hline
    DeepLab-LargeFOV& \textbf{0.21} (GPU) & 62.3\\ \hline
    DeepLab-CRF-LargeFOV & 2.13(GPU+CPU)\footnotemark & 67.6\\ \hline   
    Piecewise & 1.50(CPU) & 70.3 \\ \hline
    CRFasRNN & 0.70(GPU) & 69.6 \\ \hline
    DeepParsing & 0.30(GPU)\footnotemark & 67.8 \\ \hline
    \hline
    baseline FCN & \textbf{0.21} (GPU) & 63.4 \\ \hline
    H-ReNet &  0.24(GPU) \footnotemark & 71.1 \\ \hline
    H-ReNet + DenseCRF (2 iterations) & 0.46(GPU) & \textbf{72.6} \\ \hline
    \end{tabular}
}
\end{center}
\caption{Comparisons of testing time and labeling accuracy on the \textit{validation} set of VOC12. The prediction time of \cite{long2014fully,mostajabi2014feedforward,noh2015learning} are not reported.}
\label{tab: speed_comp_voc12}
\vspace{-3em}
\end{table}

\addtocounter{footnote}{-2}
\footnotetext{The FCN and mean field inference are implemented on GPU and CPU, respectively.}
\addtocounter{footnote}{1}
\footnotetext{In \cite{liu2015semantic}, the authors only report the time cost of the last 4 layers in their 15-layer net, which is 0.075 seconds. By building up a net, which has the exactly same 11 layers as in the DPN net, we estimate the time cost of first 11 layers to be 0.22 seconds.}
\addtocounter{footnote}{1}
\footnotetext{Time cost is measured on an image of size $375\times 500$.}

When extra training data from MS COCO are used, \textit{DeepParsing} marginally outperforms the H-ReNet ($77.5\%$ v.s. $76.8\%$). A plausible explanation is that the mixture of label contexts and high-order relations modeled by the \textit{DeepParsing} are better captured when a sufficiently large amount of training data are available.


\begin{figure}[t!]
\begin{center}
\includegraphics[width=1.05\linewidth]{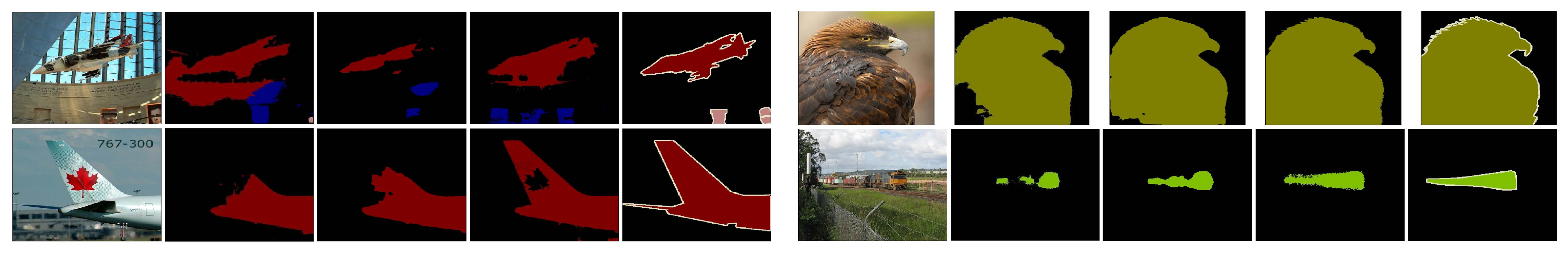}
\end{center}
\vspace{-1.em}
   \caption{Qualitative comparisons with other approaches on VOC 12 \textit{validation} set. For each example, results of input, \textit{CRFasRNN}\cite{zheng2015conditional}, \textit{DeepParsing}\cite{liu2015semantic}, H-ReNet w/ \textit{DenseCRF} and groundtruth are shown from left to right.}
\label{fig:comp_other_methods}
\vspace{-2em}
\end{figure}

\noindent \textbf{Qualitative comparisons}. In Figure \ref{fig:comp_other_methods}, we qualitatively compare the H-ReNet to two leading approaches, \textit{CRFasRNN}~\cite{zheng2015conditional} and \textit{DeepParsing}~\cite{liu2015semantic}, both with graphical models incorporated in an end-to-end manner. 
In the first example, a part of the airplane is visually similar to the background wall at the bottom. \textit{DeepParsing} mistakenly labels the background wall as the airplane while \textit{CRFasRNN} method incorrectly classifies the airplane part as the background. In contrast, the H-ReNet can recognize a more complete body of the airplane due to the explicit modeling of contextual dependence. In the second example, the part of bird torso in the lower left corner can only be recognized by the H-ReNet. In the third and fourth examples, more complete foreground regions are labeled by the H-ReNet. In general, by seeing boarder areas, the H-ReNet is more capable of resolving ambiguous regions.

\vspace{-2em}
\begin{table}[h!]
\begin{center}
\resizebox{1.05\textwidth}{!} {

    \begin{tabular}{ P{3.8cm} | P{.62cm}  P{.62cm} P{.62cm}   P{0.62cm}   P{.62cm}  P{0.62cm}  P{.62cm}  P{.62cm}  P{.62cm}  P{.62cm}  P{.62cm}  P{.62cm}  P{.68cm}   P{.88cm}   P{.88cm}   P{.84cm}   P{.88cm}   P{.82cm}   P{.62cm}   P{.65cm}  | P{.95cm}}
    \multicolumn{21}{c}{ (a) Per-class results on VOC12 test set}\\ \hline
	   Method   & aero & bike & bird & boat & bottle & bus & car & cat & chair & cow & table & dog & horse & mbike & person & plant & sheep & sofa & train & tv & IoU($\%$)\\ \hline    
    
    MSRA-CFM & 75.7 & 26.7 & 69.5 & 48.8 & 65.6 & 81.0 & 69.2 & 73.3 & 30.0 & 68.7 & 51.5 & 69.1 & 68.1 & 71.7 & 67.5 & 50.4 & 66.5 & 44.4 & 58.9 & 53.5 & 61.8\\ \hline
    FCN-8s & 76.8 & 34.2 & 68.9 & 49.4 & 60.3 & 75.3 & 74.7 & 77.6 & 21.4 & 62.5 & 46.8 & 71.8 & 63.9 & 76.5 & 73.9 & 45.2 & 72.4 & 37.4 & 70.9 & 55.1 & 62.2 \\ \hline
     
    Zoomed-out& 85.6 & 37.3 & \textbf{83.2} & 62.5 & 66.0 & 85.1 & 80.7 & 84.9 & 27.2 & 73.2 & 57.5 & 78.1 & 79.2 & 81.1 & 77.1 & 53.6 & 74.0 & 49.2 & 71.7 & 63.3 & 69.6 \\ \hline
    
    DeepLab-CRF-LargeFOV & 83.5 & 36.6 & 82.5 & 62.3 & 66.5 & 85.4 & 78.5 & 83.7 & 30.4 & 72.9 & 60.4 & 78.5 & 75.5 & 82.1 & 79.7 & 58.2 & 82.0 & 48.8 & 73.7 & 63.3 & 70.3 \\ \hline
    
    DeconvNet+ CRF\cite{noh2015learning} & 87.8 & 41.9 & 80.6 & 63.9 & 67.3 & 88.1 & 78.4 & 81.3 & 25.9 & 73.7 & 61.2 & 72.0 & 77.0 & 79.9 & 78.7 & 59.5 & 78.3 & 55.0 & 75.2 & 61.5 & 70.5 \\ \hline

    Piecewise & 87.5 & 37.7 & 75.8 & 57.4 & \textbf{72.3} & 
    88.4 & 82.6 & 80.0 & \textbf{33.4} & 71.5 & 
    55.0 & 79.3 & 78.4 & 81.3 & 82.7 & 
    56.1 & 79.8 & 48.6 & 77.1 & 66.3 & 
    70.7\\ \hline
    
    CRFasRNN & 87.5 & 39.0 & 79.7 & 64.2 & 68.3 & 87.6 & 80.8 & 84.4 & 30.4 & 78.2 & 60.4 & 80.5 & 77.8 & 83.1 & 80.6 & 59.5 & 82.8 & 47.8 & 78.3 & \textbf{67.1} & 72.0\\ \hline 
      
    DeepParsing & 87.7 & \textbf{59.4} & 78.4 & 64.9 & 70.3 & 
    89.3 & \textbf{83.5} & 86.1 & 31.7 & 79.9 & 
    \textbf{62.6} & 81.9 & 80.0 & 83.5 & 82.3 & 
    60.5 & 83.2 & 53.4 & 77.9 & 65.0 & 
    74.1 \\ \hline
    \hline
    
    H-ReNet & 86.9 &  38.6&  80.8 &  63.9 &  67.6&
			        88.5&  82.6&  87.6 &  29.3 &  79.4&  58.6  &
			        82.6 &  80.8&  83.3&  81.7&  58.7&  82.9&
			        55.7 &  79.2&  64.6  & 72.7\\ \hline  
                    
    H-ReNet+ DenseCRF & \textbf{89.7} &  39.6&  82.9 &  \textbf{65.1} &  68.9&
			        \textbf{89.8}&  83.1&  \textbf{90.1} &  29.9 &  \textbf{81.7}&  59.1  &
			        \textbf{85.1} &  \textbf{82.9}&  \textbf{84.7}&  \textbf{83.5}&  \textbf{61.5}&  \textbf{84.8}&
			        \textbf{57.8} &  \textbf{80.1}&  65.7  & \textbf{74.3}\\ \hline
    
    \multicolumn{21}{c}{ (b) Per-class results on VOC12 test set with the extra MS COCO training data }\\ \hline  
 	WSSL\cite{papandreou2015weakly} & 88.5 & 35.9 & 88.5 & 62.3 & 68.0 & 87.0 & 81.0 & 86.8 & 32.2 & 80.8 & 60.4 & 81.1 & 81.1 & 83.5 & 81.7 & 55.1 & 84.6 & 57.2 & 75.7 & 67.2 & 73.0 \\ \hline
    BoxSup\cite{dai2015boxsup} & 89.8 & 38.0 & 89.2 & \textbf{68.9} & 68.0 
    & 89.6 & 83.0 & 87.7 & 34.4 & 83.6 
    & 67.1 & 81.5 & 83.7 & 85.2 & 83.5 
    & 58.6 & 84.9 & 55.8 & 81.2 & \textbf{70.7} 
    & 75.2\\ \hline
    CRFasRNN &  90.4 & 55.3 & 88.7 & 68.4 & 69.8 & 88.3 & 82.4 & 85.1 & 32.6 & 78.5 & 64.4 & 79.6 & 81.9 & 86.4 & 81.8 & 58.6 & 82.4 & 53.5 & 77.4 & 70.1 & 74.7 \\ \hline
    DeepParsing & 89.0 & \textbf{61.6} & 87.7 & 66.8 & \textbf{74.7} 
    & \textbf{91.2} & \textbf{84.3} & 87.6 & 36.5 & \textbf{86.3} 
    & \textbf{66.1} & 84.4 & 87.8 & 85.6 & 85.4 
    & 63.6 & 87.3 & \textbf{61.3} & 79.4 & 66.4 
    & \textbf{77.5} \\ \hline
    \hline
    H-ReNet & 90.4 &  39.7&  89.8 & 65.2 & 70.4 &
			        88.7&  82.8 & 87.7 &  33.6 &  84.3&  
                    63.8  & 84.3 & 86.9&  85.7&  84.9&  
                    60.1&  86.4& 55.1 &  80.3 &  69.1  
                    & 75.3 \\ \hline 
                    
    H-ReNet+ DenseCRF & \textbf{93.1} & 42.2 &  \textbf{92.2} & 66.4 & 71.3&
			        90.2&  83.4&  \textbf{89.9} &  \textbf{36.7} &  85.2 &  
                    64.8  & \textbf{86.4} &  \textbf{88.5}&  \textbf{86.9} &  \textbf{86.3}&  
                    \textbf{62.4}&  \textbf{87.6} & 56.8 &  \textbf{81.7}&  69.9  
                    & 76.8 \\ \hline                    
    \end{tabular}   
}
\end{center}
\caption{Quantitative comparisons with external approaches on \textit{VOC12} \textit{test} set. 
}
\label{tab: results_on_voc}
\vspace{-4em}
\end{table}

\vspace{-0.5em}
\section{Conclusions}
\vspace{-0.5em}

In this work, we propose the use of the spatially recurrent layer (ReNet layer) for semantic segmentation, and demonstrate its effectiveness by constructing a naive deep ReNet (N-ReNet) and evaluting it on the \textit{Stanford Background} dataset. Furthermore, we integrate ReNet layers with FCNs and develop a hybrid network (H-ReNet), and achieve state-of-the-art results on the PASCAL VOC 2012 dataset.


\clearpage

\bibliographystyle{splncs}
\bibliography{egbib}

\begin{thebibliography}{10}

\bibitem{lin2015efficient}
Lin, G., Shen, C., Reid, I.,  et~al.:
\newblock Efficient piecewise training of deep structured models for semantic
  segmentation.
\newblock arXiv preprint arXiv:1504.01013 (2015)

\bibitem{zheng2015conditional}
Zheng, S., Jayasumana, S., Romera-Paredes, B., Vineet, V., Su, Z., Du, D.,
  Huang, C., Torr, P.:
\newblock Conditional random fields as recurrent neural networks.
\newblock In: ICCV. (2015)

\bibitem{liu2015semantic}
Liu, Z., Li, X., Luo, P., Loy, C.C., Tang, X.:
\newblock Semantic image segmentation via deep parsing network.
\newblock In: ICCV.
\newblock (2015)

\bibitem{he2015deep}
He, K., Zhang, X., Ren, S., Sun, J.:
\newblock Deep residual learning for image recognition.
\newblock arXiv preprint arXiv:1512.03385 (2015)

\bibitem{yan2015hd}
Yan, Z., Zhang, H., Piramuthu, R., Jagadeesh, V., DeCoste, D., Di, W., Yu, Y.:
\newblock Hd-cnn: Hierarchical deep convolutional neural networks for large
  scale visual recognition.
\newblock In: ICCV. (2015)

\bibitem{ren2015faster}
Ren, S., He, K., Girshick, R., Sun, J.:
\newblock Faster r-cnn: Towards real-time object detection with region proposal
  networks.
\newblock In: Advances in Neural Information Processing Systems. (2015)  91--99

\bibitem{long2014fully}
Long, J., Shelhamer, E., Darrell, T.:
\newblock Fully convolutional networks for semantic segmentation.
\newblock In: CVPR.
\newblock (2015)

\bibitem{graves2013generating}
Graves, A.:
\newblock Generating sequences with recurrent neural networks.
\newblock arXiv preprint arXiv:1308.0850 (2013)

\bibitem{graves2014towards}
Graves, A., Jaitly, N.:
\newblock Towards end-to-end speech recognition with recurrent neural networks.
\newblock In: Proceedings of the 31st International Conference on Machine
  Learning (ICML-14). (2014)  1764--1772

\bibitem{tighe2013finding}
Tighe, J., Lazebnik, S.:
\newblock Finding things: Image parsing with regions and per-exemplar
  detectors.
\newblock In: CVPR. (2013)  3001--3008

\bibitem{Tighe:2010:SSN}
Tighe, J., Lazebnik, S.:
\newblock Superparsing: Scalable nonparametric image parsing with superpixels.
\newblock In: ECCV. (2010)

\bibitem{LiuYT11}
Liu, C., Yuen, J., Torralba, A.:
\newblock Nonparametric scene parsing via label transfer.
\newblock TPAMI (2011)

\bibitem{singh2013nonparametric}
Singh, G., Kosecka, J.:
\newblock Nonparametric scene parsing with adaptive feature relevance and
  semantic context.
\newblock In: CVPR. (2013)

\bibitem{eigen2012nonparametric}
Eigen, D., Fergus, R.:
\newblock Nonparametric image parsing using adaptive neighbor sets.
\newblock In: CVPR. (2012)

\bibitem{russell2009associative}
Russell, C., Kohli, P., Torr, P.:
\newblock Associative hierarchical crfs for object class image segmentation.
\newblock In: ICCV. (2009)

\bibitem{kohli2009robust}
Kohli, P., Torr, P.H.:
\newblock Robust higher order potentials for enforcing label consistency.
\newblock IJCV (2009)

\bibitem{li2013exploring}
Li, Y., Tarlow, D., Zemel, R.:
\newblock Exploring compositional high order pattern potentials for structured
  output learning.
\newblock In: CVPR. (2013)

\bibitem{ramalingam2008exact}
Ramalingam, S., Kohli, P., Alahari, K., Torr, P.H.:
\newblock Exact inference in multi-label crfs with higher order cliques.
\newblock In: CVPR. (2008)

\bibitem{zitnick2014edge}
Zitnick, C.L., Doll{\'a}r, P.:
\newblock Edge boxes: Locating object proposals from edges.
\newblock In: Computer Vision--ECCV 2014.
\newblock Springer (2014)  391--405

\bibitem{dai2014convolutional}
Dai, J., He, K., Sun, J.:
\newblock Convolutional feature masking for joint object and stuff
  segmentation.
\newblock In: CVPR.
\newblock (2015)

\bibitem{dai2015boxsup}
Dai, J., He, K., Sun, J.:
\newblock Boxsup: Exploiting bounding boxes to supervise convolutional networks
  for semantic segmentation.
\newblock In: ICCV. (2015)

\bibitem{hariharan2014hypercolumns}
Hariharan, B., Arbel{\'a}ez, P., Girshick, R., Malik, J.:
\newblock Hypercolumns for object segmentation and fine-grained localization.
\newblock In: CVPR.
\newblock (2015)

\bibitem{hariharan2014simultaneous}
Hariharan, B., Arbel{\'a}ez, P., Girshick, R., Malik, J.:
\newblock Simultaneous detection and segmentation.
\newblock In: ECCV.
\newblock Springer (2014)  297--312

\bibitem{noh2015learning}
Noh, H., Hong, S., Han, B.:
\newblock Learning deconvolution network for semantic segmentation.
\newblock In: ICCV.
\newblock (2015)

\bibitem{girshickregion}
Girshick, R., Donahue, J., Darrell, T., Malik, J.:
\newblock Region-based convolutional networks for accurate object detection and
  segmentation.
\newblock TPAMI (2015)

\bibitem{pinheiro2013recurrent}
Pinheiro, P.H., Collobert, R.:
\newblock Recurrent convolutional neural networks for scene parsing.
\newblock JMLR (2013)

\bibitem{chen2014semantic}
Chen, L.C., Papandreou, G., Kokkinos, I., Murphy, K., Yuille, A.L.:
\newblock Semantic image segmentation with deep convolutional nets and fully
  connected crfs.
\newblock ICLR (2015)

\bibitem{schwing2015fully}
Schwing, A.G., Urtasun, R.:
\newblock Fully connected deep structured networks.
\newblock arXiv preprint arXiv:1503.02351 (2015)

\bibitem{mostajabi2014feedforward}
Mostajabi, M., Yadollahpour, P., Shakhnarovich, G.:
\newblock Feedforward semantic segmentation with zoom-out features.
\newblock In: CVPR. (2015)

\bibitem{bell2015inside}
Bell, S., Zitnick, C.L., Bala, K., Girshick, R.:
\newblock Inside-outside net: Detecting objects in context with skip pooling
  and recurrent neural networks.
\newblock arXiv preprint arXiv:1512.04143 (2015)

\bibitem{visin2015renet}
Visin, F., Kastner, K., Cho, K., Matteucci, M., Courville, A., Bengio, Y.:
\newblock Renet: A recurrent neural network based alternative to convolutional
  networks.
\newblock arXiv preprint arXiv:1505.00393 (2015)

\bibitem{zaremba2014recurrent}
Zaremba, W., Sutskever, I., Vinyals, O.:
\newblock Recurrent neural network regularization.
\newblock arXiv preprint arXiv:1409.2329 (2014)

\bibitem{chung2015gated}
Chung, J., Gulcehre, C., Cho, K., Bengio, Y.:
\newblock Gated feedback recurrent neural networks.
\newblock arXiv preprint arXiv:1502.02367 (2015)

\bibitem{simonyan2014very}
Simonyan, K., Zisserman, A.:
\newblock Very deep convolutional networks for large-scale image recognition.
\newblock CoRR (2014)

\bibitem{szegedy2014going}
Szegedy, C., Liu, W., Jia, Y., Sermanet, P., Reed, S., Anguelov, D., Erhan, D.,
  Vanhoucke, V., Rabinovich, A.:
\newblock Going deeper with convolutions.
\newblock arXiv preprint arXiv:1409.4842 (2014)

\bibitem{Jia13caffe}
Jia, Y.:
\newblock {Caffe}: An open source convolutional architecture for fast feature
  embedding (2013)

\bibitem{zeiler2014visualizing}
Zeiler, M.D., Fergus, R.:
\newblock Visualizing and understanding convolutional networks.
\newblock In: ECCV.
\newblock (2014)

\bibitem{farabet2012scene}
Farabet, C., Couprie, C., Najman, L., LeCun, Y.:
\newblock Scene parsing with multiscale feature learning, purity trees, and
  optimal covers.
\newblock In: ICML. (2012)

\bibitem{kekecc2014contextually}
Keke{\c{c}}, T., Emonet, R., Fromont, E., Tr{\'e}meau, A., Wolf, C.:
\newblock Contextually constrained deep networks for scene labeling.
\newblock In: BMVC. (2014)

\bibitem{sharma2014recursive}
Sharma, A., Tuzel, O., Liu, M.:
\newblock Recursive context propagation network for semantic scene labeling.
\newblock In: NIPS. (2014)

\bibitem{byeon2015scene}
Byeon, W., Breuel, T.M., Raue, F., Liwicki, M.:
\newblock Scene labeling with lstm recurrent neural networks.
\newblock In: CVPR. (2015)

\bibitem{arbelaez2011contour}
Arbelaez, P., Maire, M., Fowlkes, C., Malik, J.:
\newblock Contour detection and hierarchical image segmentation.
\newblock TPAMI (2011)

\bibitem{hariharan2011semantic}
Hariharan, B., Arbel{\'a}ez, P., Bourdev, L., Maji, S., Malik, J.:
\newblock Semantic contours from inverse detectors.
\newblock In: ICCV. (2011)

\bibitem{everingham2010pascal}
Everingham, M., Van~Gool, L., Williams, C.K., Winn, J., Zisserman, A.:
\newblock The pascal visual object classes (voc) challenge.
\newblock IJCV (2010)

\bibitem{lin2014microsoft}
Lin, T.Y., Maire, M., Belongie, S., Hays, J., Perona, P., Ramanan, D.,
  Doll{\'a}r, P., Zitnick, C.L.:
\newblock Microsoft coco: Common objects in context.
\newblock In: ECCV.
\newblock (2014)

\bibitem{yu2015multi}
Yu, F., Koltun, V.:
\newblock Multi-scale context aggregation by dilated convolutions.
\newblock arXiv preprint arXiv:1511.07122 (2015)

\bibitem{ioffe2015batch}
Ioffe, S., Szegedy, C.:
\newblock Batch normalization: Accelerating deep network training by reducing
  internal covariate shift.
\newblock arXiv preprint arXiv:1502.03167 (2015)

\bibitem{papandreou2015weakly}
Papandreou, G., Chen, L.C., Murphy, K., Yuille, A.L.:
\newblock Weakly-and semi-supervised learning of a dcnn for semantic image
  segmentation.
\newblock arXiv preprint arXiv:1502.02734 (2015)

\end{thebibliography}
\end{document}